\documentclass[runningheads]{llncs}

\usepackage[year=2024,ID=12234]{eccv}

\usepackage{eccvabbrv}

\usepackage{graphicx}
\usepackage{booktabs}
\usepackage{tabularx}
\usepackage{multirow}
\usepackage{xcolor,colortbl}
\usepackage[pagebackref,breaklinks,colorlinks]{hyperref}
\usepackage{comment}
\usepackage[accsupp]{axessibility}  

\definecolor{first_color}{HTML}{FFBFBF}
\definecolor{second_color}{HTML}{FFFBCC}
\definecolor{third_color}{HTML}{FFDFBF}
\newcommand{\first}[1]{{\cellcolor{first_color} #1}}
\newcommand{\second}[1]{{\cellcolor{second_color} #1}}

\definecolor{title1}{HTML}{DA627D}
\definecolor{title2}{HTML}{ff8655}
\definecolor{title3}{HTML}{539EC7}
\definecolor{title4}{HTML}{9A348E}

\usepackage{orcidlink}
\usepackage{booktabs}
\usepackage{framed}

\usepackage{floatrow}
\DeclareFloatFont{tiny}{\tiny}
\begin{document}

\title{\textcolor{title1}{Snap-it}, \textcolor{title2}{Tap-it}, \textcolor{title3}{Splat-it}: Tactile-Informed 3D Gaussian Splatting for Reconstructing Challenging Surfaces} 

\titlerunning{Tactile-Informed 3D Gaussian Splatting}

\author{Mauro~Comi\inst{1} \and Alessio~Tonioni\inst{2} \and Max~Yang\inst{1} \and Jonathan~Tremblay\inst{3} \and Valts~Blukis\inst{3} \and Yijiong~Lin\inst{1} \and Nathan~F.~Lepora$^*$\inst{1} \and Laurence~Aitchison$^*$\inst{1}}

\authorrunning{Comi et al.}



\institute{$^1$University of Bristol, $^2$Google Zurich, $^3$NVIDIA}

\maketitle

\begin{figure}[ht!]
\centerline{\includegraphics[width=1\textwidth]{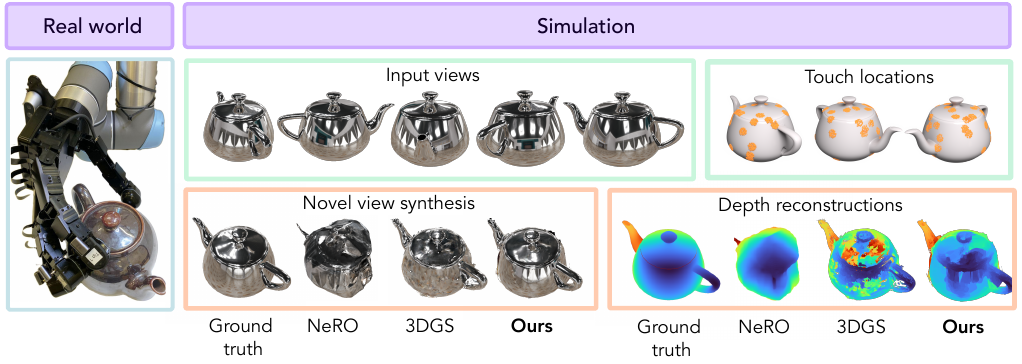}}
\caption{
We combine multi-view visual data and tactile sensing information within a 3D Gaussian Splatting framework 
for accurate geometry reconstruction and novel view synthesis of challenging surfaces. 
On the left, a robotic arm is equipped with tactile sensors. 
On the right, we show results on novel view synthesis and depth reconstruction, using a five minimal-view setting and tactile input (first row) -- 
touches not only improve surface reconstruction, but also strengthen novel view synthesis.
}
  \label{fig:fig1}
  
\end{figure}

\vspace{-8mm}

\begin{abstract}
Touch and vision go hand in hand, mutually enhancing our ability to understand the world.
From a research perspective, the problem of mixing touch and vision together is underexplored and presents 
interesting challenges.
To this end, we propose \textit{Tactile-Informed 3DGS}, a novel approach that incorporates touch data (local depth maps) with multi-view vision data to achieve surface reconstruction and novel view synthesis. 
Our method optimises 3D Gaussian primitives to accurately model the object's geometry at points of contact. 
By creating a framework that decreases the transmittance at touch locations, 
we achieve a refined surface reconstruction, 
ensuring a uniformly smooth depth map.
Touch is particularly useful when considering non-Lambertian objects (\textit{e.g.} shiny or reflective surfaces) since contemporary methods tend to fail to reconstruct with fidelity specular highlights.   
By combining vision and tactile sensing, we achieve more accurate geometry reconstructions with fewer images than prior methods. 
We conduct evaluation on objects with glossy and reflective surfaces and demonstrate the effectiveness of our approach, offering significant improvements in reconstruction quality.
\end{abstract}

\section{Introduction}
\label{sec:intro}
Humans perceive the world through a multitude of senses, and often, visual perception alone is not sufficient to fully grasp the intricacy of the world.
When only using vision, it is possible to misunderstand shapes, materials, scale, \textit{etc.} of the world and its components. Tactile sensing, however, provides consistent geometric information that complements visual perception, particularly in scenarios where visual data may be insufficient. Thus, in this paper we are interested in exploring the problem of combining two sources of sensing, vision and touch, for novel view synthesis and shape reconstruction from a set of views and touches (see Figure~\ref{fig:fig1}). While non trivial, the task of combining vision and touch for accurate 3D reconstruction presents numerous advantages. Current vision-based methods often struggle with \textit{non-Lambertian surfaces} \cite{mildenhall2021nerf}, as well as scenarios where only a \textit{limited number of views} are available. These scenarios are common in real-world settings, such as robotics manipulation, VR/AR, and 3D modeling, where physical constraints or occlusions can restrict the number of accessible viewpoints. Unlike visual data, tactile sensing provides consistent geometric information regardless of light-dependent effects, and having data from an additional modality can support 3D reconstruction in the context of both limited (5 views) and dense visual data.
Conversely, while tactile sensing provides valuable geometric information, it is limited by the fact that it only captures partial information about an object's surface - {\em e.g.}, when grasping an object, the hand rarely touches the entire surface.
Therefore, the ability to seamlessly combine sparse observations across modalities and scenarios with varying viewpoint availability is what we aim to achieve in this work.

We believe to be the first to explore the problem of solely using RGB and 3D touches 
to reconstruct an object and allow for novel view synthesis. 
Prior work has looked at the problem of mixing vision and touch signals~\cite{smith2020,smith2021} for point cloud retrieval. Other methods have focused on reconstructing object shapes using touch signals~\cite{qi2023general,comi2023touchsdf}.
While these methods tend to generate valuable object surfaces, it is rare for them to 
perform novel view synthesis from touch only; and even if they do, generating novel views is not their main objective ~\cite{suresh2023neural}. 
Ultimately, these methods are applied to Lambertian objects, and recently, vision-only based methods 
have shown tremendous success in novel view synthesis and 3D reconstructions for this setting~\cite{mildenhall2021nerf,kerbl3Dgaussians}.
These advancements are quite astonishing; however, because they leverage volumetric rendering, 
their ability to reconstruct surfaces is limited, especially in the presence of non-Lambertian surfaces. 
Moreover, depth sensors tend to fail in this setting as the highly reflective surfaces can create discontinuities in depth measurements~\cite{tao2015depth}. 
As such, the problem of recovering textured non-Lambertian objects from multiple views has been studied recently leveraging inverse rendering~\cite{liu2023nero}. 
This approach can produce very accurate results when the method is allowed an extensive amount of views and unbounded compute time (24+ hours) \cite{liu2023nero, verbin2022ref}. 

Motivated by Liu \textit{et al.}'s~\cite{liu2023nero} research on non-Lambertian 3D reconstructions, we propose Tactile-Informed 3DGS, a novel multimodal interaction approach that integrates tactile sensing and vision within 3D Gaussian Splatting for challenging object reconstruction. Differently from prior work, we streamlined this problem by reducing the required number of viewpoints and computation time through the incorporation of sparse 3D touch observations on the object's surface. Unlike visual data, tactile sensing is not affected by the presence of specular highlights, reflections, and translucenct materials.
A key insight is that, by regularising the 3D transmittance of Gaussians around the touch locations, our approach directly guides the optimisation procedure with precise, direct information regarding object geometry. 
Acknowledging the sparse nature of tactile observations, we further refine our method by introducing an unsupervised regularisation term to smooth the predicted camera depth maps.

By leveraging multimodal sensing, Tactile-Informed 3DGS provides three main contributions: (1) state-of-the-art geometry reconstruction on reflective and glossy surfaces, (2) faster scene reconstruction 10x over prior arts, and (3) improved performance in geometry reconstruction and novel view synthesis with minimal views.
We evaluate our method on object-centric datasets containing glossy and reflective surfaces, namely Shiny Blender \cite{verbin2022ref} and Glossy Synthetic \cite{liu2023nero} datasets. We also propose a real-world dataset of a highly shiny object (metallic toaster) from which we capture a multiview set of images as well as real robot touches. 
Our results show that multimodal sensing yields equal or better geometry reconstruction than methods that incorporate notion of light scattering, while still being an order of magnitude faster. Moreover, by introducing touch as an additional sensing modality, our method increases robustness and mitigates performance degradation when working with fewer viewpoints.

\section{Literature Review}
\subsection{Tactile sensing for 3D object reconstruction}
Tactile sensing for object reconstruction is an emerging research field that integrates techniques from computer vision, graphics, and robotics with high-resolution, optical-based tactile sensors \cite{yuan2017gelsight, lambeta2020, lepora2021}. The study of affordable and open-source tactile sensors has contributed to advancements in 3D shape reconstruction through direct object interactions. 
These sensors, which capture local depth maps or marker-based images, can be leveraged to collect information about the surface at touch location. 

Recently we have seen the introduction of datasets that incorporate touch, such as ObjectFolder~\cite{gao2023objectfolder}.
This dataset, however, does not include multiview images of the objects and it does not include non-Lambertian objects, whereas our method focuses on mixing touch and vision. As such, we propose a dataset that includes multiple views of a metallic object.
Suresh \textit{et al.}~\cite{Suresh22icra} introduced a dataset that both mixes an external depth sensor and touch sensing of objects, although they do not provide a vision-based multiview system.  

Wang et al. \cite{wang20183d,yuan2017gelsight} proposed to utilise tactile exploration for object reconstruction through voxel-based representations. 
Smith et al. \cite{smith2020, smith2021} introduced a novel approach to decouple vision and tactile sensing processes, employing a Convolutional Neural Network (CNN) to map tactile readings into local contact surface representations. 
The reconstruction was achieved by combining geometrical features extracted with tactile sensors with feature embeddings collected using RGB cameras.
Their experiments focused mainly on synthetic data containing objects with simple material proprieties (single color and only Lambertian materials) and in the end, their method does not allow for novel view synthesis.

\subsection{Novel-view synthesis on reflective surfaces}

Novel-view synthesis (NVS) from sparse observations is a key area of research within the domain of computer graphics, 3D computer vision, and robotics. 
Methods based on Neural Radiance Fields (NeRF) are extremely successful at synthesizing photorealistic images by modeling the volumetric density and color of light rays within a scene with a Multi-Layer Perceptron (MLP)~\cite{mildenhall2021nerf, muller2022instant, barron2021mip, park2021nerfies}. 
3D Gaussian Splatting (3DGS)~\cite{kerbl3Dgaussians} has been proposed as an alternative to NeRF. 
Instead of modelling the radiance field implicitly through a neural network, it is modelled explicitly by a large set of 3D Gaussians. 
This has the benefit of a simpler explicit model that offers faster training and rendering than even highly optimized NeRF implementations~\cite{muller2022instant}. 
In this work we propose to extend the 3D Gaussian splatting framework to reason about non-Lambertian surfaces
and touch data. 

The application of volumetric rendering equations causes difficulties for both NeRFs and 3DGS in accurately representing specular and glossy surfaces. This is because it is challenging to interpolate the outgoing radiance across different viewpoints as it tends to vary significantly around specular highlights \cite{verbin2022ref}. Ref-NeRF\cite{verbin2022ref} was proposed to address this limitation by replacing NeRF's view-dependent radiance parameterisation with a representation of reflected radiance, leveraging spatially varying scene properties and a regulariser on normal vectors. 
%
NeRFReN~\cite{guo2022nerfren} models reflections by fitting an additional radiance field for the reflected world beyond the mirror, however this method supports reflections on planar surfaces only.
Ref-NeRF focuses on direct light, thus struggling with indirect lights from nearby light emitters. 
NeRO \cite{liu2023nero}, on the other hand, proposes to model direct and indirect illumination separately using the split-sum approximation \cite{karis2013real}. 
This method divides the equation for surface reflectance into two components: one that pre-integrates the lighting and another that integrates the characteristics of the material. 
By directly approximating indirect specular lighting, NeRO achieves improved geometry reconstruction.
The results they obtained for material retrieval and surface reconstruction are impressive, 
but as we show in this work, when the method does not have access to an extensive dataset of views, 
it tends to generate unconvincing results (see experiments section). 

\section{Preliminaries}
\label{sec:related_work}

\subsection{3D Gaussian Splatting}
3D Gaussian Splatting \cite{kerbl3Dgaussians} is an explicit radiance field technique for modeling 3D scenes as a large set of 3D Gaussians. Specifically, a scene is parameterised as a set of $N$ anisotropic 3D Gaussians, each defined by a mean vector $\boldsymbol{\mu} \in \mathbb{R}^3$, a covariance matrix $\mathbf{\Sigma} \in \mathbb{R}^3$, an opacity $\alpha \in \mathbb{R}^1$ , and a view-dependent color vector $\mathbf{c} \in \mathbb{R}^m$ encoded as spherical harmonics. These parameters are optimised during the training process through backpropagation. The influence of a Gaussian on a point $\mathbf{x} \in \mathbb{R}^3$ is determined by:

\begin{equation}
f(\mathbf{x};\mathbf{\Sigma}, \boldsymbol{\mu}) = \exp(-\frac{1}{2} (\mathbf{x} - \boldsymbol{\mu})^T \mathbf{\Sigma}^{-1} (\mathbf{x} - \boldsymbol{\mu})).
\end{equation}

During the optimisation process, the scene is rendered to a 2D plane by calculating each Gaussian's projected covariance $\mathbf{\Sigma'} \in \mathbb{R}^2$ and position $\boldsymbol{\mu}' \in \mathbb{R}^2$, following:

\begin{equation}
\mathbf{\Sigma'} = \mathbf{J} \mathbf{W} \mathbf{\Sigma} \mathbf{W}^T \mathbf{J}^T,
\end{equation}

where $\mathbf{W}$ is the projective transformation and $\mathbf{J}$ is the Jacobian of the affine transformation. The covariance matrix $\mathbf{\Sigma}$, along with other parameters, is optimised to ensure a positive semi-definite matrix through decomposition:

\begin{equation}
\mathbf{\Sigma} = \mathbf{R}\mathbf{S}\mathbf{S}^T\mathbf{R}^T.
\end{equation}

The 2D mean vector $\boldsymbol{\mu}'$ results from perspective projection $\text{Proj}(\boldsymbol{\mu} | \mathbf{E}, \mathbf{K})$, with $\mathbf{E}$ and $\mathbf{K}$ being the camera's extrinsic and intrinsic matrices. Consequently, the pixel color $I(\mathbf{p})$ in the image space is computed by summing the contributions of all Gaussians, and accounting for their color, opacity, and influence:

\begin{equation}
I(p) = \sum_{i=1}^{N} f^{2D}(\textbf{p}; \boldsymbol{\mu'}_i, \mathbf{\Sigma'}_i) \ \alpha_i \ \mathbf{c}_{i} \prod_{j=1}^{i-1} 1 - f^{2D}(\textbf{p}; \boldsymbol{\mu'}_{j}, \mathbf{\Sigma'}_{j}) \ \alpha_{j}
\end{equation}

Similarly, the depth is computed as:
\begin{equation}
\label{eq:depth}
D(p) = \sum_{i=1}^{N} f^{2D}(\textbf{p}; \boldsymbol{\mu'}_i, \mathbf{\Sigma'}_i) \ \alpha_i \ d_{i} \prod_{j=1}^{i-1} 1 - f^{2D}(\textbf{p}; \boldsymbol{\mu'}_{j}, \mathbf{\Sigma'}_{j}) \ \alpha_{j}
\end{equation}

where $d_i$ is the depth of the $i\text{-th}$ Gaussian obtained by projecting $\boldsymbol{\mu}_i$ onto the image frame. 

The optimisation process starts by initialising the 3D Gaussians using Structure-from-Motion (SfM) point clouds. It then proceeds to optimise the learnable parameters by minimising the photometric loss between predicted and ground truth images. During the entire procedure, Gaussians are adaptively densified or pruned based on heuristics on their opacity and mean gradient changes.

\section{Methodology}
\label{sec:methodology}

This section outlines our approach to reconstructing 3D objects by integrating visual and tactile data through 3D Gaussian Splatting. Our methodology follows two primary stages. The first stage involves the generation of initial point clouds from local depth maps, while the second step consists of optimising and regularising a Gaussian representation to accurately model the object's surface.

\subsection{Gaussian initialisation and optimisation}
The optimisation process starts with the extraction of a pointcloud from COLMAP, where each point's position and associated color serve as the initial mean and color attributes for a \textit{first} set of Gaussians $G_c$. The view-dependent color is encoded using spherical-harmonics, which are essential in modeling the specular and glossy highlights. This initial point cloud is combined with a further set of points $P_t$ collected with an optical tactile sensor, on which a \textit{second} set of Gaussians $G_t$ is initialised on tactile data. Specifically, each point in $P_t$ serves as the mean for one of the  Gaussians in $G_t$, while their color and scale attributes are initialised randomly. A uniform opacity value is assigned across all Gaussians in both $G_t$ and $G_c$. Both sets of Gaussians can be cloned, split, and removed according to the heuristics outlined by Kerbl et al. \cite{kerbl3Dgaussians}. The newly generated Gaussians that result from this process are assigned to the same set of their originals. 

The optimisation of these Gaussians involves minimising the photometric loss between predicted and ground truth RGB image. This loss is further regularised by the edge-aware smoothness loss described in \ref{sec:smooth_aware}. In addition to the photometric and edge-aware smoothness loss, Gaussians $G_t$ are regularised by the 3D transmittance loss, which we discuss in Sect. \ref{sec:transmittance_loss}.

\subsection{Regularisation}

\subsubsection{3D transmittance}
\label{sec:transmittance_loss} The transmittance quantifies the degree to which light penetrates through a medium. Given $N$ Gaussians and a point $\mathbf{p} \in \mathbb{R}^3$, where $\mathbf{p}$ belongs to the touch surface, we define the average 3D transmittance at $\mathbf{p}$ as:

\begin{equation}
\hat{T}(\mathbf{p}) = \dfrac{1}{N} \sum_{j=1}^{N} 1 - f(\textbf{p}; \boldsymbol{\mu}_{j}, \mathbf{\Sigma}_{j}) \ \alpha_{j}
\end{equation}

\noindent where $\boldsymbol{\mu}_i \in \mathbb{R}^3$ and $\mathbf{\Sigma_i} \in \mathbb{R}^3$ are the mean and covariance of the unprojected 3D Gaussians. 
To manage the computational load and prioritise the optimisation of the Gaussians around the touch locations, we limit the number of Gaussians considered per point to those with the highest spatial influence on $p$. 
Moreover, we incorporate a distance-based filtering criterion designed to exclude Gaussians beyond a certain threshold distance from $\mathbf{p}$. 
This is because, in the process of selecting Gaussians based on their spatial impact, it is possible to include Gaussians that, while influential, are positioned at considerable distances from the target point. Given that the Gaussian rasteriser subdivides the frustum into discrete tiles, thus calculating a point's transmittance solely from Gaussians within the same tile, distant Gaussians — even those with potential impact — would not account for the final transmittance at $\mathbf{p}$. We define our loss $\mathcal{L}_T$ as the average over all the $\mathbf{p} \in P_t$ points collected on the object surfaces:

\begin{equation}
\label{eq:lt}
\mathcal{L}_T = - \dfrac{1}{P} \sum_{i=1}^P \hat{T}(\mathbf{p_i})
\end{equation}
Minimising the average transmittance around the touch location effectively brings the Gaussians closer to the touched surfaces and increases their opacity, thereby implicitly forcing them to model the real underlying surface. 

\begin{figure}[t!]
    \centerline{\includegraphics[width=1\textwidth]{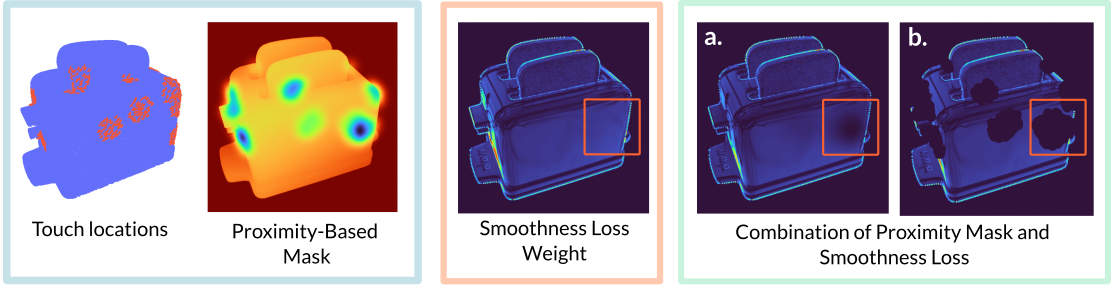}}
    \caption{Proximity-Based Mask's impact on the gradients computed by the edge-aware smoothness loss. \textit{Left}: points collected on the object's surface alongside the derived proximity mask. \textit{Centre}: averaged horizontal and vertical gradients as determined by the smoothness loss, where lighter shades correspond to higher gradients. \textit{Right}: integration of the smoothness loss with the proximity mask using two distinct approaches: \textbf{(a)} implementation of a distance-based Gaussian decay within the proximity mask, and \textbf{(b)} masking based on a discrete threshold from the contact surface.}
    \label{fig:prox_mask}
\end{figure}

\subsubsection{Edge-Aware Smoothness with Proximity-Based Masking}
\label{sec:smooth_aware}
Given the sparsity of touch readings, we leverage an edge-aware smoothness loss $\mathcal{L}_S$ \cite{heise2013pm} to refine the surface reconstruction further. This regularisation is based on the intuition that depth transitions within an image are often aligned with intensity changes in the RGB space. The edge-aware smoothness function defined as:

\begin{equation}
\label{eq:lt}
\mathcal{L}_S = \frac{1}{N} \sum_{i, j} \left( |\partial_x D_{i, j}| e^{-\beta |\partial_x I_{i, j}|} + |\partial_y D_{i, j}| e^{-\beta |\partial_y I_{i, j}|} \right)
\end{equation}
where $D$ is the rasterised depth map and $I$ is the ground truth image for a given view. The horizontal and vertical gradients are calculated by applying convolution operations with 5x5 Sobel kernels \cite{vairalkar2012edge}. In contrast to the smaller kernel size commonly adopted by edge detectors, the filters we adopt capture more contextual information, leading to a smoother gradient estimation and thus less instability around specular highlights. 

While the edge-aware smoothness loss positively contributes to the refinement of surface reconstruction, its integration with the transmittance loss presents challenges. Specifically, the smoothness loss interferes with the transmittance loss around touch locations — areas where we aim to prioritise transmittance loss to accurately reconstruct contact surfaces. This is because the smoothness loss operates on a global scale, affecting the continuity and appearance of the entire scene or large parts of it. In contrast, the transmittance loss has a more localised effect, focusing on areas around the known points. When the smoothness loss acts on areas close to the known geometry, it may counteract the localized enhancements intended by the transmittance loss - for example, by minimising opacity in order to decrease gradients in the depth image. To address this, we introduce a \textit{Proximity-Based Masking} strategy. This approach involves generating a proximity mask via a Gaussian decay function, centered around the projected 3D touch locations. This mask serves to delineate regions of higher confidence based on depth information and visibility, thereby modulating the edge-aware smoothness loss according to the proximity to touch locations. 

The creation of a proximity-based mask is achieved by calculating the 2D Euclidean distance transform \cite{fabbri20082d} between every pixel in the rasterised depth map and its nearest projected 3D point. Then, a Gaussian decay is applied based on the calculated distance, producing the actual proximity mask. Finally, the edge-aware smoothness loss, which aims to align the gradients of the predicted depth map with those of the corresponding RGB image, 
is modulated by the inverse of this proximity mask, effectively diminishing the loss's influence in areas close to the contact surfaces and preserving edge details as the distance from touch locations increases. The integration of transmittance loss with our edge-aware smoothness loss with proximity-based masking ensures a comprehensive reconstruction quality: while the transmittance loss facilitates the reconstruction of the contact surfaces, the smoothness loss refines the overall surface by mitigating inaccuracies in less explored regions.

\section{Results}
\subsection{Datasets \& Evaluation}
We train and evaluate our method on two object-centric datasets containing glossy and reflective surface: Shiny Blender \cite{verbin2022ref} and Glossy Reflective \cite{liu2023nero}. 
The Shiny Blender \cite{verbin2022ref} datasets consists of 6 objects characterised by non-Lambertian materials. For each object, the dataset provides 100 training images and 200 test images all accompanied by  camera poses. 
The \textit{ball} object was excluded in our evaluation due to the absence of ground truth depth maps, 
which makes the assessment of the reconstructed geometry infeasible. 
Although ground truth depth maps and normal maps are available, 
our model relies solely on RGB images and local depth maps collected on the object surface by simulating a touch sensor, 
and we never leverage ground truth depth maps for supervision. 

The Glossy Synthetic \cite{liu2023nero} datasets consist of 8 objects, each represented by 128 images at a resolution of 800x800. The dataset is divided into 32 images for training and 96 images for testing. This dataset ranges from objects with large smooth surfaces displaying specular effects (e.g., Bell, Cat) to those with complex geometries (e.g., Angel, Luyu).
 
 The evaluation metric employed to asses geometry reconstruction is the Chamfer Distance (CD)\cite{sun2018pix3d}, which measures the similarity between predicted and ground truth point clouds. 
The Glossy Synthetic dataset provides the ground truth point cloud for CD calculation. 
For the Shiny Blender dataset, we create ground truth point clouds for CD calculation by projecting the ground truth depth maps in the three-dimensional space. 
This method ensures that we only consider areas visible to the camera, thereby accounting for occlusions and ensuring a fair comparison. 
The decision against using point clouds sampled directly from the ground truth meshes is aimed at avoiding discrepancies caused by occluded regions which are not comparable across different methods. 
For the predicted point clouds, we apply the method adopted on the Glossy Synthetic dataset. 
In addition, we provide a qualitative comparison  between standard 3DGS and our proposed method on a dataset collected in the real-world (Sect. \ref{sec:real_world}).

\subsection{Glossy Synthetic}

\setlength\tabcolsep{0.1em}

\begin{table}[tb]
   \caption{Evaluation of the Chamfer Distance (CD) on the Glossy Synthetic dataset \cite{liu2023nero} dataset. Our method is the best in recovering the object geometry from glossy surfaces, followed by NeRO.}
        \begin{tabular}{c c c c c c c c c c}
        \toprule
        \multicolumn{10}{c}{Glossy Synthetic \cite{liu2023nero}, 100 views} \\
        
        Method &  Angel &  Bell & Cat &  Horse &  Luyu & Potion & Tbell &  Teapot & Avg. \\
                
        \midrule
        \multicolumn{10}{c}{CD($\downarrow$)} \\
        \midrule
        Ref-NeRF & 0.0291 & 0.0137 & 0.0201  & 0.0071 & 0.0141 & 0.0131 & 0.0216 & 0.0143  & 0.0241 \\
        NeRO & \second 0.0034 & \first  0.0032 & \second 0.0044 & 0.0049 & 0.0054 & \second 0.0053 & \first  0.0035 & \first  0.0037 & \second 0.0042 \\
        3DGS &  \first  0.0002 &   0.0132 &   0.0073 &  \second 0.0005 & \second  0.0011 &   0.0143 &   0.0133 &   0.0098 &   0.0075 \\
         \textbf{Ours}[5 grasps] &   \first 0.0002 &   \second 0.0068 &   \first  0.0027 & \first 0.0004  & \first   0.0009 &   \first  0.0038 & \second 0.0072 & \second  0.0059 &  \first  0.0034 \\
         
        \bottomrule

        \end{tabular}
    \vspace{0.3em}

   \label{table:glossy_full}
\end{table}

\noindent\textbf{Full-views}: Table \ref{table:glossy_full} benchmarks our method against 3D Gaussian Splatting (3DGS), NeRO, and Ref-NeRF in the settings where all methods have full access to all the 100 training views. We chose these methods as they represent the state-of-the-art in terms of quality reconstruction for challenging objects. Additionally, our method uses 5 grasps per object, whereas a grasp consists of ``five'' fingers touching the object, resulting in a total of 25 tactile readings. 
To obtain the results of 3D Gaussian Splatting we trained using the official 3DGS implementation, while for NeRO and Ref-NeRF we directly report the available results from \cite{liu2023nero}, although perceptual metrics (PSNR and SSIM) are not provided. To compute the CD for 3D Gaussian Splatting-related methods, the predicted point cloud is derived from projecting rasterised depth maps to 3D, using the 96 test cameras. This results in an average of 500K points per object. 

Our analysis demonstrates that our method significantly outperforms the leading competitor, 3DGS, in 3D reconstruction quality. With just five grasps, our approach achieves a 50\% decrease in the average Chamfer Distance (CD), resulting in 0.0034 compared to 3DGS's 0.0075. This performance places our 3DGS-based method on par with, but in some cases ahead of, current NeRF-based state-of-the-art methods such as NeRO and Ref-NeRF. Morevoer, radiance field approaches such as NeRO and Ref-NeRF are considerably more computationally demanding. For reference, while NeRO requires 25 hours of training \cite{liu2023nero}, our method takes 1 hour on average to reconstruct an object with 5 grasps. 

\setlength\tabcolsep{0.2em}
\begin{table*}[h]
\caption{Evaluation of the Chamfer Distance (CD) and PSNR on 5-views on the Glossy Synthetic dataset \cite{liu2023nero}. In a minimal view setting, our method performs better than 3DGS and NeRO both in terms of geometry reconstruction and novel-view synthesis.}

  \label{table:glossy_minimal}
    \centering
        \begin{tabular}{c c c c c c c c c c}
        \toprule
        \multicolumn{10}{c}{Glossy Synthetic \cite{liu2023nero}, 5 views} \\
        
        Method &  Angel &  Bell & Cat &  Horse &  Luyu & Potion &      Tbell &  Teapot & Avg. \\
                
        \midrule
        \multicolumn{10}{c}{CD($\downarrow$)} \\
        \midrule
   3DGS & \second 0.0004 & \second 0.0220 & 0.0295 & \second 0.0007 & \second 0.0013 & 0.0122 & \second 0.0145 & \second 0.0079 & \second 0.0111 \\
    NeRO & 0.0893 & 0.0398 & \second 0.0230 & 0.1817 & 0.0170 & \second 0.0043 & 0.0859 & 0.0282 & 0.0586 \\

     \textbf{Ours}[5 grasps] & \first 0.0003	&  \first 0.0082	& \first 0.0021 & \first 0.0004 & \first 0.0008 & \first 0.0013 & \first 0.0057 & \first 0.0022 & \first 0.0026 \\        
       \midrule        
        
        \multicolumn{10}{c}{PSNR($\uparrow$)} \\
        \midrule
   3DGS & \second 20.98 & \first 19.60 & \second 22.77 & \second 18.99 & \first 21.24 & \second 21.33 & \second 17.97 & \second 17.80 & \second	20.08  \\

    NeRO & 10.93 & 17.33 & 15.53 & 9.76 & 13.58  & 17.44 & 12.54 & 13.13 & 13.78  \\

     \textbf{Ours}[5 grasps] & \first 21.02 & \second 19.52 & \first 23.06 & \first 21.94 & \second 20.95 &	\first 21.64 & \first  18.25 & \first 17.94 & \first 20.55  \\        
        \bottomrule

        \end{tabular}

\end{table*}

\begin{figure}[ht!]
\centerline{\includegraphics[width=1\textwidth]{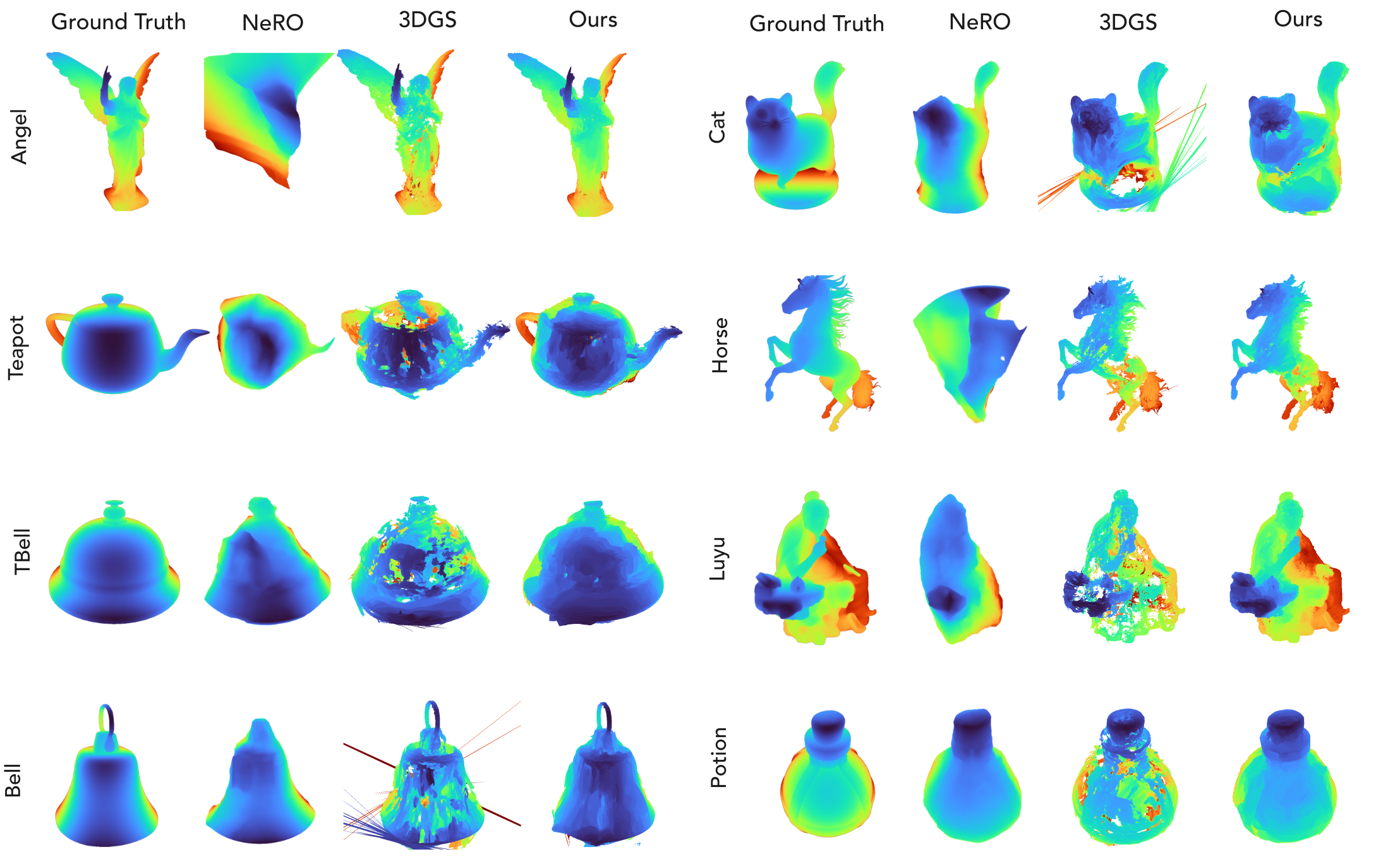}}
\caption{Surface reconstruction qualitative results from 5 training views on the Glossy Synthetic dataset}
  \label{fig:qualitative}
\end{figure}

\begin{figure}[ht!]
\centerline{\includegraphics[width=1\textwidth]{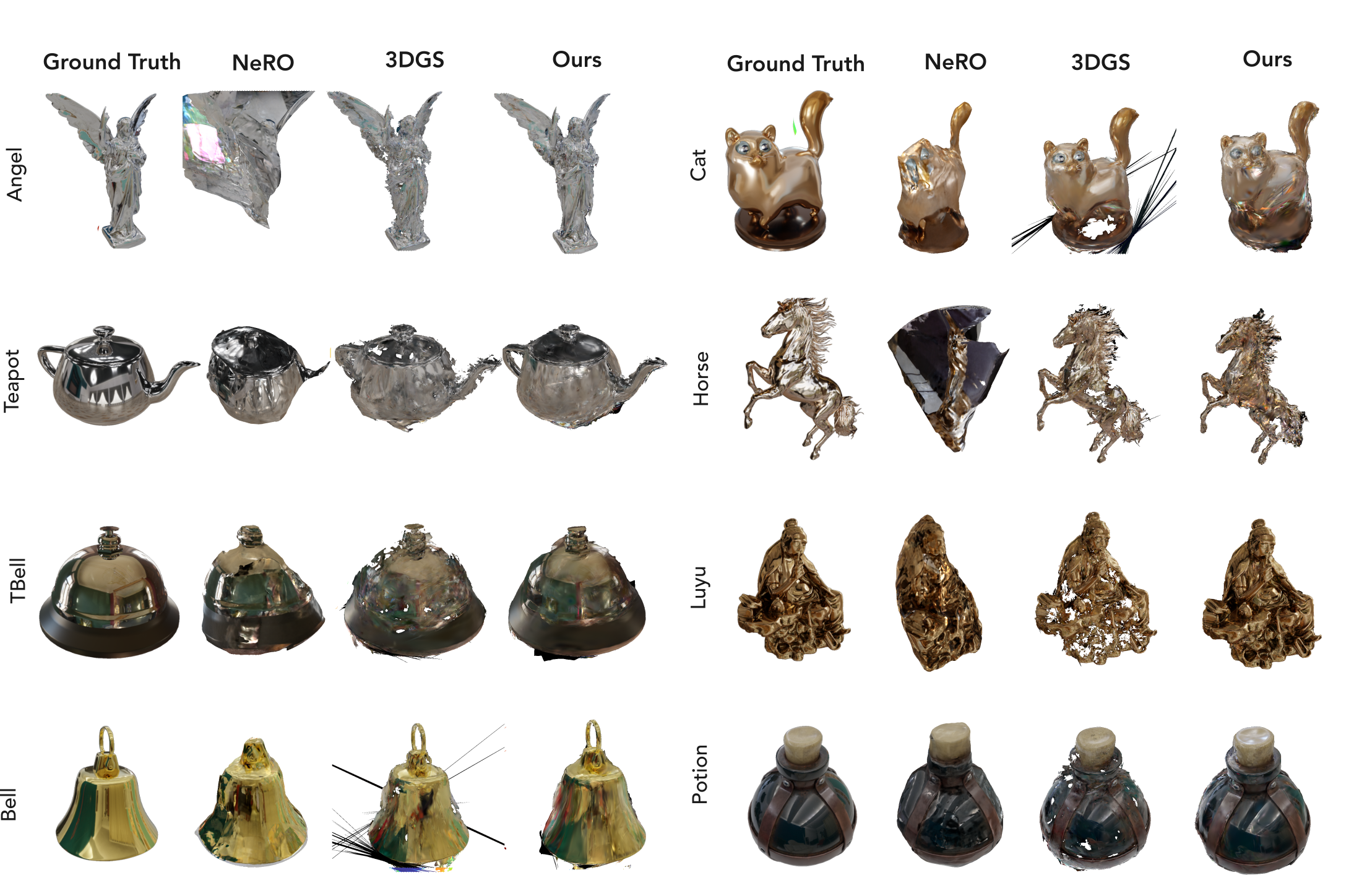}}
\caption{Novel-view synthesis qualitative results from 5 training views on the Glossy Synthetic dataset}
  \label{fig:qualitative_rgb}
\end{figure}

\noindent\textbf{Minimal-views}: We explore the performance of the different methods in a ``minimal views'' regime where models are provided with only 5 input views. In this setting we evaluate the top 3 performing methods from the previous experiment: 3DGS, NeRO and Ours. We used NeRO's official implementation to retrain the model. 
After training, we computed the CD for NeRO models as described in the original paper and extracted meshes using the Marching Cubes algorithm \cite{lorensen1998marching}. To ensure consistency in evaluation, we sampled around 500,000 points on the generated meshes, aligning with the previously detailed evaluation procedure for 3DGS. Figure~\ref{fig:qualitative} and~\ref{fig:qualitative_rgb} show that in this setting the addition of grasps significantly improves the quality of the reconstructed geometry and multiview rendering.  
Table \ref{table:glossy_minimal} reports quantitative results and shows how NeRO and 3DGS quality of 3D reconstruction drops dramatically when trained with few views, while our method can retain and, for some models, even improve. In this setting, our method achieves over 4 times improvement over the second best model in terms of geometry accuracy (0.0026 Ours vs 0.0111 3DGS), clearly showing the effectiveness of our formulation for high quality 3D reconstruction. The improved reconstruction contributes to enhanced quality in novel view synthesis, as demonstrated by the PSNR scores.

\setlength\tabcolsep{0.6em}
\begin{table}[h!]
   \caption{Evaluation of SSIM, PSNR, and CD results on the Shiny Blender dataset \cite{verbin2022ref}. Our full method includes both touches and our proposed smoothness loss. Our method considerably improves the geometry reconstruction, while achieving comparable levels of image fidelity.}
   \label{tab:shiny_blender}

    \centering
        \begin{tabular}{c c c c c c c}
        \toprule
        \multicolumn{7}{c}{Shiny Blender\cite{verbin2022ref}} \\
        
        Method &  Car &  Coffee & Helmet &     Teapot &    Toaster & Avg. \\
        
        \midrule
        \multicolumn{7}{c}{CD($\downarrow$)} \\
        \midrule
       
        3DGS &   0.0027 &    0.0018 &   0.0068 &    \second 0.0003 &    0.0069 &  0.0037 \\
        
         3DGS + S & \second 0.0014 &   0.0022 &   \second 0.0024 &   0.0007 &   0.0077 &  0.0028 \\
         
         3DGS + T[5 grasps] & 0.0028 &	\first 0.0011	& 0.0043& \first	0.0002	& \first 0.0029 & \second 0.0022 \\
         
         \textbf{Ours}[5 grasps] &  \first 0.0004 & \second 0.0017 & \first  0.0005 & \first  0.0002 & \second 0.0038 & \first  0.0013 \\

        \midrule
        \multicolumn{7}{c}{PSNR($\uparrow$)} \\
        \midrule
        
        3DGS &  27.40 &   32.81 &   27.56 & \second  45.48 &    21.10 &  30.87 \\
        
         3DGS + S & \first  27.46 & \second  32.90 &  \first 27.69 &   45.45 &    21.12 &  \first 30.92 \\

         3DGS + T[5 grasps] & 27.36 & 32.88 & \second 27.60 & \first 45.51 & \second 21.16 & \second 30.90 \\
         
         \textbf{Ours}[5 grasps] & \second 27.43 &  \first  32.91 &   27.48 &   45.10 &    \first 21.19 &  30.82 \\

        \midrule
        \multicolumn{7}{c}{SSIM($\uparrow$)} \\
        \midrule

        3DGS &   \second 0.930 &  \first  0.972 &   0.950 &  \first 0.997 &    0.896 &  \second 0.949 \\
        
         3DGS + S &  \first 0.932 &  \first 0.972 &  \first 0.952 & \first  0.997 &  \second  0.898 & \first  0.950 \\

         3DGS + T[5 grasps] & \second 0.930 & \second 0.971 & 0.948 & \second 0.996 & 0.896	& 0.948 \\
         
         \textbf{Ours}[5 grasps] & \first 0.932 &   \first 0.972 &  \second 0.951 &  \first 0.997 & \first  0.899 &  \first 0.950 \\
         
    \bottomrule
        \end{tabular}
    \vspace{0.3em}

\end{table}

\subsection{Shiny Blender}
We continue the evaluation on the Shiny Blender dataset~\cite{verbin2022ref} where we focus on methods based on 3D Gaussian Splatting. 
We are interested in exploring the impact of the different design choices and as such we compare our method against 
a baseline defined by vanilla 3DGS, a  method that only uses the edge-aware smoothness regularisation introduced in Section \ref{sec:smooth_aware} (3DGS+S), and a method that only uses 3DGS and tactile information (3DGS + T).

\begin{figure}[t!]
\centerline{\includegraphics[width=1\textwidth]{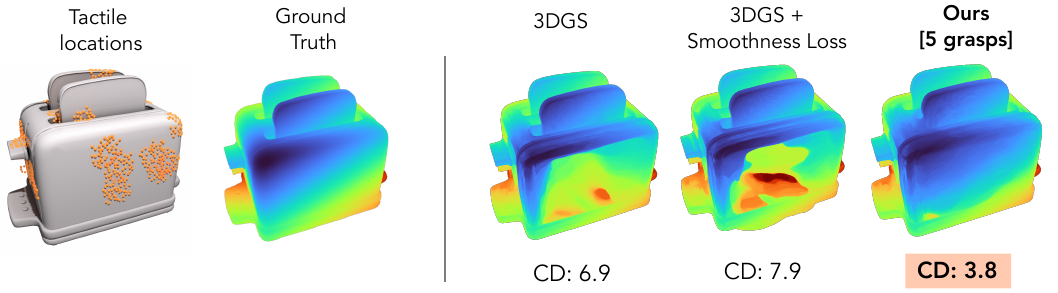}}
\caption{Rasterised depth maps of the \textit{Toaster} object. Our method results in a smoother, accurate reconstruction compared to 3DGS and 3DGS with additional regularisation on the smoothness loss.}
\label{fig:toaster}
\end{figure}

Table \ref{tab:shiny_blender} reports quantitative results comparing the four variants of Gaussian Splatting-based methods according to both the quality of the 3D reconstruction (CD) and of novel view synthesis (PSNR and SSIM).
Our method does not sacrifice the quality of the rendering as shown by neutral metrics in terms of PSNR and SSIM compared to 3DGS, while it significantly improves the geometry of the reconstructions as measured by Avg. CD (1.3 Ours vs 3.72 3DGS). 
The gap is more pronounced on models with relatively large shiny surfaces like \textit{toaster} and \textit{helmet}. 
Please also refer to Figure~\ref{fig:toaster} for qualitative visualisations of the difference in reconstructions. 
The results of 3DGS+S also show how the smoothness loss is effective to improve the quality of 3D reconstruction, but it still falls short compared to our full method where constraints given by the surfaces reconstructed by the grasps and the smoothness regularisation act in synergy to further boost performance. Similarly, while the addition of touches in 3DGS+T results in better reconstruction quality over the standalone 3DGS model, the absence of smoothness regularisation limits its ability to refine the reconstruction beyond the areas of contact.  

\begin{figure}[t!]
\centerline{\includegraphics[width=0.9\textwidth]{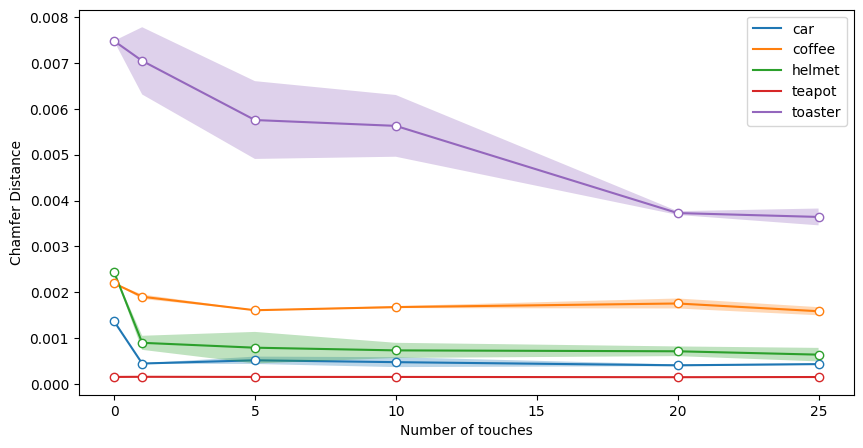}}
\caption{Correlation between the number of touches and its impact on surface reconstruction for the Glossy Synthetic dataset. 
For this experiment we kept the number of views constant to the full training dataset. 
}
\label{fig:nb_touches}
\end{figure}

We study the Chamfer Distance (CD) as a function of the number of touches used for initialising and optimising the Gaussians in our scene representation (Figure~\ref{fig:nb_touches}). 
Our results reveal consistent improvement in reconstruction accuracy with an increase in physical interactions with the objects. Given the stochastic nature of our sampling process, we conducted five independent runs per number of touches to ensure the reliability of our results. 
This validation highlights the robustness of our findings, confirming a trend of improved reconstruction quality up to a point of convergence.  

\begin{figure}[t!]
\centerline{\includegraphics[width=0.95\textwidth]{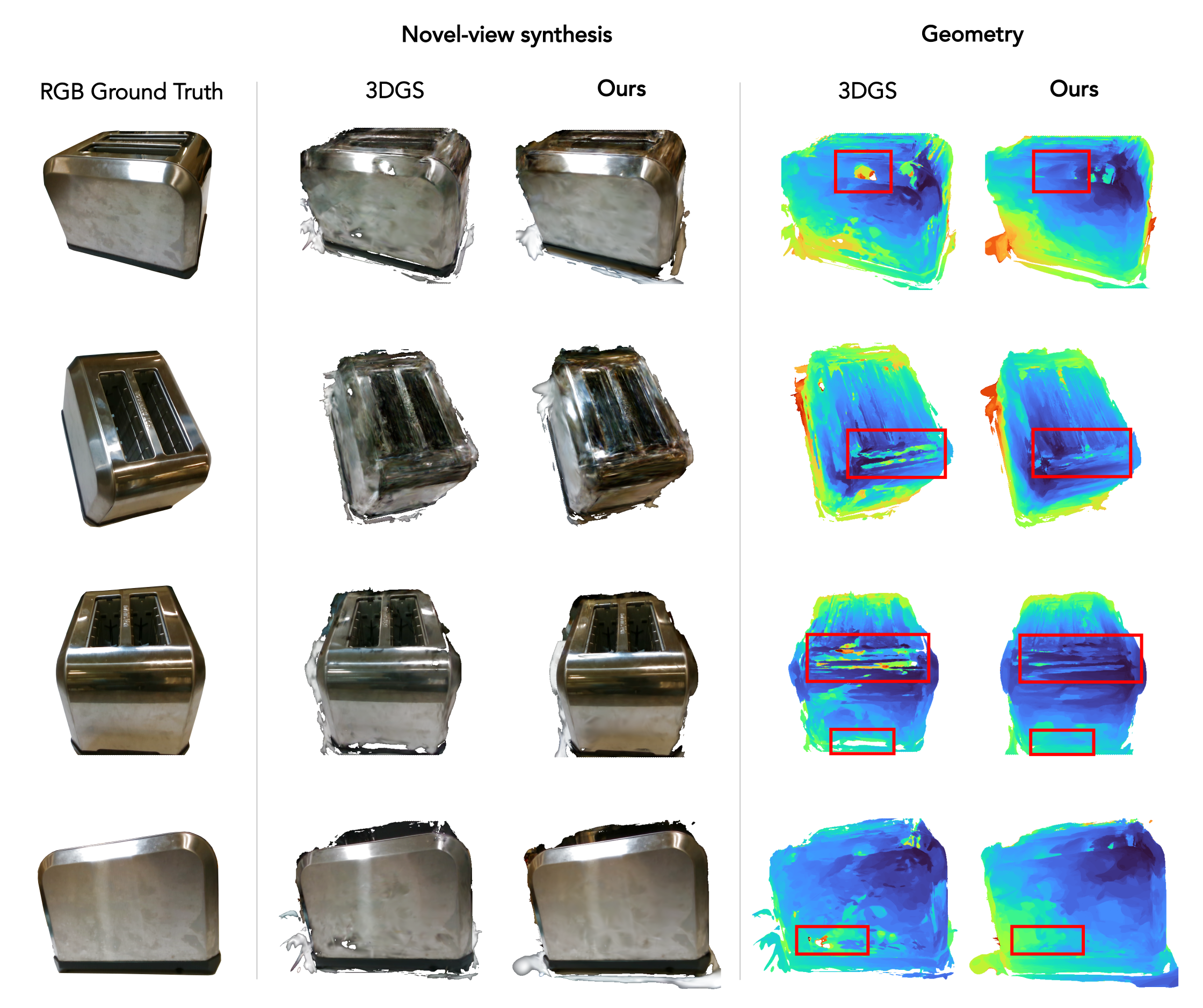}}
\caption{Qualitative reconstructions in the real-world. Highlighted areas show that the introduction of touches leads to smoother and more accurate surface reconstruction.}
  \label{fig:real_comparison}
\end{figure}

\subsection{Real world experiment}
\label{sec:real_world}
In addition to the experiments in simulated environments, we conduct a qualitative comparison on real world reconstruction. We employed a robotic hand (Allegro Hand) equipped with four tactile sensors and an in-hand camera to acquire multiple views and grasps of a real object, aiming to reconstruct its geometry using both a standard 3D Gaussian Splatting model (3DGS) and our novel approach. Specifically, we collected 25 views and 20 tactile readings to capture the object's geometry. The qualitative results presented here (Fig. \ref{fig:real_comparison}) are part of an evaluation to demonstrate the improvements our method offers in real-world geometric reconstruction of challenging surfaces. Due to the unavailability of ground truth geometry, the calculation of the Chamfer Distance is not feasibile. 
For clearer visualisation and to emphasise the quality of object reconstruction, we presented cropped images and 3D Gaussian representations focused on the object, though training utilised the original, uncropped images. 

Our results highlight regions where our approach produces smoother surfaces or eliminates holes. These findings align with outcomes from simulated experiments, indicating that the inclusion of local depth maps obtained from tactile sensors improves significantly reconstruction quality. For more information on the data collection procedure employed in this experiment, please refer to the supplementary material.
\label{sec:results}

\section{Limitations and Conclusion}
\label{sec:conclusion}
This work introduces a novel approach for incorporating touch sensing with vision. 
We believe to be the first to have explored the problem of reconstruction and novel view synthesis of an object that is both seen and touched.
We presented convincing results both qualitatively and quantitatively on two complex datasets, and we showed that our method outperforms baselines, especially when a set of minimal views of the object are used. 

Despite these promising results, there are some limitations that highlight areas for future research. 
The current methodology employs random touch sampling to collect contact surfaces for scene reconstruction. 
This strategy may not always be efficient and could be extended by an adaptive sampling method, which can select sampling locations to complement visual data. 
Future work could focus on investigating how multimodal interaction can further improve reconstruction of transparent objects within the scene. 
We believe incorporating ideas like surface modelling and surface normals can also improve upon our results.

\section*{Acknowledgements}
We would like to thank Alex Zook and Stan Birchfield for their feedback regarding the scope and contributions of our research,  which helped improve the quality of our work.



%
%
\bibliographystyle{splncs04}
\bibliography{egbib}
\end{document}